\pdfoutput=1

\documentclass[11pt,a4paper]{article}
\usepackage[hyperref]{acl2020}
\usepackage{times}
\usepackage{latexsym}

\usepackage{microtype}

\usepackage{url}

\usepackage{graphicx}
\usepackage{multirow}
\usepackage{multicol}

\aclfinalcopy
\def\aclpaperid{12}

\title{Lipschitz Constrained Parameter Initialization for Deep Transformers}

\author{
Hongfei Xu$^{1,2}$\ \ \ \ Qiuhui Liu$^3$\thanks{\ \ \ \ Corresponding author.}\ \ \ \ Josef van Genabith$^{1,2}$\ \ \ \ Deyi Xiong$^{4}$\ \ \ \ Jingyi Zhang$^{2}$\\
$^1$Saarland University / Saarland, Germany\\
$^2$German Research Center for Artificial Intelligence / Saarland, Germany\\
$^3$China Mobile Online Services / Henan, China\\
$^4$Tianjin University / Tianjin, China\\
\{hfxunlp, liuqhano\}@foxmail.com,
Josef.Van\_Genabith@dfki.de,\\
dyxiong@tju.edu.cn,
Jingyi.Zhang@dfki.de
}
\date{}

\begin{document}
\maketitle
\begin{abstract}
The Transformer translation model employs residual connection and layer normalization to ease the optimization difficulties caused by its multi-layer encoder/decoder structure. Previous research shows that even with residual connection and layer normalization, deep Transformers still have difficulty in training, and particularly Transformer models with more than 12 encoder/decoder layers fail to converge. In this paper, we first empirically demonstrate that a simple modification made in the official implementation, which changes the computation order of residual connection and layer normalization, can significantly ease the optimization of deep Transformers. We then compare the subtle differences in computation order in considerable detail, and present a parameter initialization method that leverages the Lipschitz constraint on the initialization of Transformer parameters that effectively ensures training convergence. In contrast to findings in previous research we further demonstrate that with Lipschitz parameter initialization, deep Transformers with the \emph{original} computation order can converge, and obtain significant BLEU improvements with up to 24 layers. In contrast to previous research which focuses on deep encoders, our approach additionally enables Transformers to also benefit from deep decoders.
\end{abstract}

\section{Introduction}

Neural machine translation has achieved great success in the last few years \cite{bahdanau2014neural,gehring2017convolutional,vaswani2017attention}. The Transformer \cite{vaswani2017attention}, which has outperformed previous RNN/CNN based translation models \cite{bahdanau2014neural,gehring2017convolutional}, is based on multi-layer self-attention networks and can be trained very efficiently. The multi-layer structure allows the Transformer to model complicated functions. Increasing the depth of models can increase their capacity but may also cause optimization difficulties \cite{mhaskar2016learning,telgarsky2016benefits,eldan2016power,he2016deep,bapna2018training}. In order to ease optimization, the Transformer employs residual connection and layer normalization techniques which have been proven useful in reducing optimization difficulties of deep neural networks for various tasks \cite{he2016deep,ba2016layer}.

However, even with residual connections and layer normalization, deep Transformers are still hard to train: the original Transformer \cite{vaswani2017attention} only contains 6 encoder/decoder layers. \newcite{bapna2018training} show that Transformer models with more than 12 encoder layers fail to converge, and propose the Transparent Attention (TA) mechanism which combines outputs of all encoder layers into a weighted encoded representation. \newcite{wang2019learning} find that deep Transformers with proper use of layer normalization are able to converge and propose to aggregate previous layers' outputs for each layer. \newcite{wu2019depth} explore incrementally increasing the depth of the Transformer Big by freezing pre-trained shallow layers. Concurrent work closest to ours is \newcite{zhang2019improving}. They address the same issue, but propose a different layer-wise initialization approach to reduce the standard deviation.

\begin{figure*}[t]
  \centering
  \includegraphics[width=2.0\columnwidth]{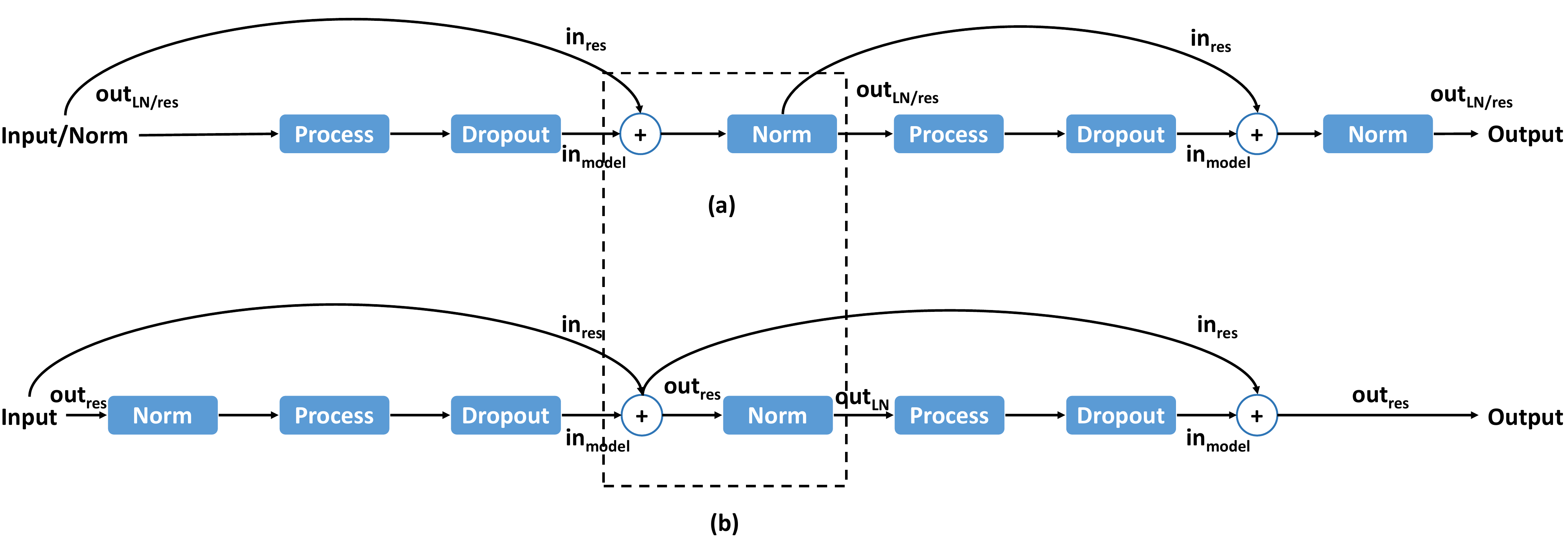}\\
  \caption{Two Computation Sequences of Transformer Translation Models: (a) the one used in the original paper, (b) the official implementation. We suggest to regard the output of layer normalization ($out_{LN/res}$) as the output of residual connection rather than the addition of $in_{res}$ and $in_{model}$ for (a), because it ($out_{LN/res}$) is the input ($in_{res}$) of the next residual connection computation.}\label{fig:vantrans}
\end{figure*}

Our contributions are as follows:
\begin{itemize}
\item We empirically demonstrate that a simple modification made in the Transformer's official implementation \cite{tensor2tensor} which changes the computation order of residual connection and layer normalization can effectively ease its optimization;
\item We deeply analyze how the subtle difference of computation order affects convergence in deep Transformers, and propose to initialize deep Transformers under the Lipschitz constraint;
\item In contrast to previous works, we empirically show that with proper parameter initialization, deep Transformers with the original computation order can converge;
\item Our simple approach effectively ensures the convergence of deep Transformers with up to 24 layers, and achieves $+1.50$ and $+0.92$ BLEU improvements over the baseline on the WMT 14 English to German task and the WMT 15 Czech to English task;
\item We further investigate deep decoders for the Transformer in addition to the deep encoders studied in previous works, and show that deep decoders can also benefit the Transformer.
\end{itemize}

\section{Convergence of Different Computation Orders}

In this paper we focus on the convergence of the training of deep transformers. To alleviate the training problem for the standard Transformer model, Layer Normalization \citep{ba2016layer} and Residual Connection \citep{he2016deep} are adopted.

\begin{table*}[t]
  \centering
    \begin{tabular}{|c|c|c|l|l|l|l|}
    \hline
    \multirow{2}[4]{*}{Models} & \multicolumn{2}{c|}{Layers} & \multicolumn{2}{c|}{En-De} & \multicolumn{2}{c|}{Cs-En} \\
\cline{2-7}          & \multicolumn{1}{c|}{Encoder} & \multicolumn{1}{c|}{Decoder} & \multicolumn{1}{c|}{v1} & \multicolumn{1}{c|}{v2} & \multicolumn{1}{c|}{v1} & \multicolumn{1}{c|}{v2} \\
    \hline
    \multicolumn{1}{|c|}{\citet{bapna2018training}$^*$} & \multicolumn{1}{c|}{16} & \multicolumn{1}{c|}{6} & \multicolumn{1}{l|}{28.39} & \multicolumn{1}{c|}{None} & \multicolumn{1}{c|}{29.36} & \multicolumn{1}{c|}{None} \\
    \cline{2-7}
    \multicolumn{1}{|c|}{\citet{wang2019learning}} & \multicolumn{1}{c|}{30} & \multicolumn{1}{c|}{6} & \multicolumn{1}{l|}{29.3} & \multicolumn{3}{c|}{{\multirow{3}[6]{*}{None}}} \\
    \cline{2-4}
    \multicolumn{1}{|c|}{\citet{wu2019depth}} & \multicolumn{2}{c|}{8} & \multicolumn{1}{l|}{29.92} & \multicolumn{3}{c|}{} \\
    \cline{2-4}
    \multicolumn{1}{|c|}{\citet{zhang2019improving}} & \multicolumn{2}{c|}{20} & \multicolumn{1}{l|}{28.67} & \multicolumn{3}{c|}{} \\
    \hline
    \multirow{4}[6]{*}{Transformer$^*$} & \multicolumn{2}{c|}{6} & \multicolumn{1}{l|}{27.77$^\ddag$} & 27.31  & \multicolumn{1}{l|}{28.62} & 28.40 \\
\cline{2-7}    & \multicolumn{2}{c|}{12} & $\neg$ & 28.12 & $\neg$ & 29.38 \\
\cline{2-7}          & \multicolumn{2}{c|}{18} & $\neg$ & 28.60 & $\neg$ & 29.61 \\
\cline{2-7}          & \multicolumn{2}{c|}{24} & $\neg$ & \textbf{29.02} & $\neg$ & \textbf{29.73}\\
    \hline
    \end{tabular}
  \caption{Results of Different Computation Orders. ``$\neg$'' means fail to converge, ``None'' means not reported in original works, ``*'' indicates our implementation of their approach. $\dag$ and $\ddag$ mean $p < 0.01$ and $p < 0.05$ while comparing between v1 (the official publication) and v2 (the official implementation) with the same number of layers in the significance test. \newcite{wu2019depth} use the Transformer Big setting, while the others are based on the Transformer Base Setting. \newcite{zhang2019improving} use merged attention decoder layers with a $50k$ batchsize.}
  \label{tab:bleuco}
\end{table*}

\begin{table*}[t]
  \centering
    \begin{tabular}{|c|c|}
    \hline
    v1 & v2 \\
    \hline
    \multicolumn{2}{|c|}{$\mu = mean(in _{model} + in _{res})$} \\
    \hline
    \multicolumn{2}{|c|}{$\sigma = std(in _{model} + in _{res})$} \\
    \hline
    \multicolumn{2}{|c|}{$out _{LN}  = \frac{{(in _{model} + in _{res} - \mu)}}{{{\sigma}}} * w + b$} \\
    \hline
    $out_{res}^{v1} = ou{t_{LN}} = \frac{w}{\sigma }*out_{res}^{v2} - \frac{{\rm{w}}}{\sigma }{\rm{*}}\mu  + b$ & $out _{res}^{v2} = in _{res} + in _{model}$ \\
    \hline
    \end{tabular}
  \caption{Computation with Layer Normalization and Residual Connection. v1 and v2 stand for the computation order of the original Transformer paper and that of the official implementation respectively. ``mean'' and ``std'' are the computation of mean value and standard deviation. $in_{model}$ and $in _{res}$ stand for output of current layer and accumulated outputs from previous layers respectively. $w$ and $b$ are weight and bias of layer normalization which are initialized with a vector full of $1$s and another vector full of $0$s. $ou{t_{LN}}$ is the computation result of the layer normalization. $out_{res}^{v1}$ and $out_{res}^{v2}$ are results of residual connections of v1 and v2.}
  \label{tab:hln}
\end{table*}

\subsection{Empirical Study of the Convergence Issue}

The official implementation \cite{tensor2tensor} of the Transformer uses a different computation order (Figure \ref{fig:vantrans} b) compared to the published version \cite{vaswani2017attention} (Figure \ref{fig:vantrans} a), since it (Figure \ref{fig:vantrans} b) seems better for harder-to-learn models.\footnote{\url{https://github.com/tensorflow/tensor2tensor/blob/v1.6.5/tensor2tensor/layers/common_hparams.py\#L110-L112}.} Even though several studies \cite{chen2018best,P18-1167} have mentioned this change and although \newcite{wang2019learning} analyze the difference between the two computation orders during back-propagation, and \newcite{zhang2019improving} point out the effects of normalization in their work, how this modification impacts on the performance of the Transformer, especially for deep Transformers, has not been deeply studied before. Here we present both empirical convergence experiments (Table \ref{tab:bleuco})  and a theoretical analysis of the effect of the interaction between layer normalization and residual connection (Table \ref{tab:hln}).

In order to compare with \newcite{bapna2018training}, we used the same datasets from the WMT 14 English to German task and the WMT 15 Czech to English task for our experiments. We applied joint Byte-Pair Encoding (BPE) \cite{sennrich2015neural} with 32k merge operations. We used the same setting as the Transformer base \cite{vaswani2017attention} except the number of warm-up steps was set to $8k$.

Parameters were initialized with Glorot Initialization\footnote{Uniformly initialize matrices between $[ - \sqrt {\frac{6}{{(isize + osize)}}} ,+ \sqrt {\frac{6}{{(isize + osize)}}} ]$, where $isize$ and $osize$ are two dimensions of the matrix.} \cite{xavier2010understanding} like in many other Transformer implementations \cite{klein2017opennmt,hieber2017sockeye,tensor2tensor}. We conducted experiments based on the Neutron implementation \citep{xu2019neutron} of the Transformer translation model. Our experiments run on 2 GTX 1080 Ti GPUs, and a batch size of $25k$ target tokens is achieved through gradient accumulation of small batches.

We used a beam size of 4 for decoding, and evaluated tokenized case-sensitive BLEU with the averaged model of the last 5 checkpoints saved with an interval of 1,500 training steps.

Results of the two different computation orders are shown in Table \ref{tab:bleuco}, which shows that deep Transformers with the computation order of the official implementation (v2) have no convergence issue.

\subsection{Theoretical Analysis}

Since the subtle change of computation order results in large differences in convergence, we further analyze the differences between the computation orders to investigate how they affect convergence.

We conjecture that the convergence issue of deep Transformers is perhaps due to the fact that layer normalization over residual connections in Figure \ref{fig:vantrans} (a) effectively reduces the impact of residual connections due to subsequent layer normalization, in order to avoid a potential explosion of combined layer outputs \cite{chen2018best}, which is also studied by \newcite{wang2019learning,zhang2019improving}. We therefore investigate how the layer normalization and the residual connection are computed in the two computation orders, shown in Table \ref{tab:hln}.

Table \ref{tab:hln} shows that the computation of residual connection in v1 is weighted by $\frac{w}{\sigma }$ compared to v2, and the residual connection of previous layers will be shrunk if $\frac{w}{\sigma } < 1.0$, which makes it difficult for deep Transformers to converge.

\section{Lipschitz Constrained Parameter Initialization}

Since the diminished residual connections (Table \ref{tab:hln}) may cause the convergence issue of deep v1 Transformers, is it possible to constrain $\frac{w}{\sigma } \ge 1.0$? Given that $w$ is initialized with $1$, we suggest that the standard deviation of $in _{model} + in _{res}$ should be constrained as follows:

\begin{equation}
	0.0 < \sigma  = std(i{n_{model}} + i{n_{res}}) \le 1.0
	\label{eqa:restrict}
\end{equation}

\noindent in which case $\frac{w}{\sigma }$ will be greater than or at least equal to $1.0$, and the residual connection of v1 will not be shrunk anymore. To achieve this goal, we can constrain elements of $i{n_{model}} + i{n_{res}}$ to be in $[a, b]$ and ensure that their standard deviation is smaller than $1.0$.

Let's define $P(x)$ as any probability distribution of $x$ between $[a, b]$:

\begin{equation}
	\int\limits_{a}^{\rm{b}} {P(x)dx = 1.0}
	\label{eqa:pl}
\end{equation}

then the standard deviation of $x$ is:

\begin{equation}
	\sigma (P(x),x) = \sqrt {\int\limits_{a}^{\rm{b}} {P(x){{\Big{(}x - \int\limits_{a}^{\rm{b}} {P(x)xdx} \Big{)}}^2}dx} }
	\label{eqa:lastequal}
\end{equation}

Given that $({x - \int\limits_{a}^{\rm{b}} {P(x)xdx} }) < (b - a)$ for ${\rm{x}} \in [a, b]$ as $P(x)$ is constrained by Equation \ref{eqa:pl}, we reformulate Equation \ref{eqa:lastequal} as follows:

\begin{equation}
	\sigma (P(x),x) < \sqrt {\int\limits_a^{\rm{b}} {P(x){{(b - a)}^2}dx} }
	\label{eqa:sig}
\end{equation}

From Equation \ref{eqa:sig} we obtain:

\begin{equation}
	\sigma (P(x),x) < (b - a)\sqrt {\int\limits_{a}^{\rm{b}} {P(x)dx} }
	\label{eqa:allmost}
\end{equation}

After applying Equation \ref{eqa:pl} in Equation \ref{eqa:allmost}, we find that:

\begin{equation}
	\sigma (P(x),x) < b - a
\end{equation}

Thus, as long as $b - a \le 1$ (the range of elements of the representation $x$), the requirements for the corresponding $\sigma$ described in Equation $1$ can be satisfied.

To achieve this goal, we can simply constrain the range of elements of $x$ to be smaller than $1$ and initialize the sub-model before layer normalization to be a k-Lipschitz function, where $k \le 1$. Because if the function $F$ of the sub-layer is a k-Lipschitz function, for inputs ${\rm{x, y }} \in {\rm{ [a, b]}}$, $|F(x)-F(y)| < k|x-y|$ holds. Given that $|x-y| \le b - a$, we can get $|F(x)-F(y)| < k(b-a)$, the range of the output of that sub-layer is constrained by making it a k-Lipschitz function with constrained input.

The k-Lipschitz constraint can be satisfied effectively through weight clipping,\footnote{Note that the weight of the layer normalization cannot be clipped, otherwise residual connections will be more heavily shrunk.} and we empirically find that deep Transformers are only hard to train at the beginning and only applying a constraint to parameter initialization is sufficient, which is more efficient and can avoid a potential risk of weight clipping on performance. \newcite{zhang2019improving} also show that decreasing parameter variance
at the initialization stage is sufficient for ensuring the convergence of deep Transformers, which is consistent with our observation.

\section{Experiments}

We use the training data described in Section 2 to examine the effectiveness of the  proposed Lipschitz constrained parameter initialization approach.

In practice, we initialize embedding matrices and weights of linear transformations with uniform distributions of $[-e, +e]$ and $[-l, +l]$ respectively. We use $\sqrt {\frac{2}{{esize + vsize}}}$ as $e$ and $\sqrt {\frac{1}{{isize}}}$ as $l$ where $esize$, $vsize$ and $isize$ stand for the size of embedding, vocabulary size and the input dimension of the linear transformation respectively.\footnote{To preserve the magnitude of the variance of the weights in the forward pass.}

\begin{table}[t]
  \centering
    \begin{tabular}{|r|l|l|l|l|}
    \hline
    \multicolumn{1}{|c|}{\multirow{2}[4]{*}{Layers}} & \multicolumn{2}{c|}{En-De} & \multicolumn{2}{c|}{Cs-En} \\
\cline{2-5}          & \multicolumn{1}{c|}{v1-L} & \multicolumn{1}{c|}{v2-L} & \multicolumn{1}{c|}{v1-L} & \multicolumn{1}{c|}{v2-L} \\
    \hline
    6     & 27.96$^\dag$  & 27.38  & 28.78$^\ddag$ & 28.39 \\
    \hline
    12    & 28.67$^\dag$  & 28.13  & 29.17 & 29.45 \\
    \hline
    18    & 29.05$^\ddag$  & 28.67  & 29.55 & 29.63 \\
    \hline
    24    & \textbf{29.46}  & 29.20  & 29.70 & \textbf{29.88} \\
    \hline
    \end{tabular}
  \caption{Results with Lipschitz Constrained Parameter Initialization.}
  \label{tab:bleunew}
\end{table}

Results for two computation orders with the new parameter initialization method are shown in Table \ref{tab:bleunew}. v1-L indicates v1 with Lipschitz constrained parameter initialization, the same for v2-L.

Table \ref{tab:bleunew} shows that deep v1-L models do not suffer from convergence problems anymore with our new parameter initialization approach. It is also worth noting that unlike \newcite{zhang2019improving}, our parameter initialization approach does not degrade the translation quality of the 6-layer Transformer, and the 12-layer Transformer with our approach already achieves performance comparable to the 20-layer Transformer in \newcite{zhang2019improving} (shown in Table \ref{tab:bleuco}).

While previous approaches \cite{bapna2018training,wang2019learning} only increase the depth of the encoder, we suggest that deep decoders should also be helpful. We analyzed the influence of deep encoders and decoders separately and results are shown in Table \ref{tab:bleued}.

\begin{table}[t]
  \centering
    \begin{tabular}{|r|r|r|r|}
    \hline
    \multicolumn{1}{|l|}{Encoder} & \multicolumn{1}{l|}{Decoder} & \multicolumn{1}{l|}{En-De} & \multicolumn{1}{l|}{Cs-En} \\
    \hline
    \multicolumn{2}{|c|}{6}     & 27.96 & 28.78 \\
    \hline
    24    & 6     & 28.76 & 29.20 \\
    \hline
    6     & 24    & 28.63 & 29.36 \\
    \hline
    \multicolumn{2}{|c|}{24}    & \textbf{29.46} & \textbf{29.70} \\
    \hline
    \end{tabular}
  \caption{Effects of Encoder and Decoder Depth with Lipschitz Constrained Parameter Initialization.}
  \label{tab:bleued}
\end{table}

Table \ref{tab:bleued} shows that the deep decoder can indeed benefit performance in addition to the deep encoder, especially on the Czech to English task.

\section{Conclusion}

In contrast to previous works \citep{bapna2018training,wang2019learning,wu2019depth} which show that deep Transformers with the computation order as in \newcite{vaswani2017attention} have difficulty in convergence, we show that deep Transformers with the original computation order can converge as long as proper parameter initialization is performed.

We first investigate convergence differences between the published Transformer \citep{vaswani2017attention} and its official implementation \citep{tensor2tensor}, and compare the differences of computation orders between them. We conjecture that the convergence issue of deep Transformers is because layer normalization sometimes shrinks residual connections, we support our conjecture with a theoretical analysis (Table \ref{tab:hln}), and propose a Lipschitz constrained parameter initialization approach for solving this problem.

Our experiments show the effectiveness of our simple approach on the convergence of deep Transformers, which achieves significant improvements on the WMT 14 English to German and the WMT 15 Czech to English news translation tasks. We also study the effects of deep decoders in addition to deep encoders extending previous works.

\section*{Acknowledgments}

We thank anonymous reviewers for their insightful comments. Hongfei Xu acknowledges the support of China Scholarship Council ([2018]3101, 201807040056). Deyi Xiong is supported by the National Natural Science Foundation of China (Grant No. 61861130364), the Natural Science Foundation of Tianjin (Grant No. 19JCZDJC31400) and the Royal Society (London) (NAF$\backslash$R1$\backslash$180122). Hongfei Xu, Josef van Genabith and Jingyi Zhang are supported by the German Federal Ministry of Education and Research (BMBF) under the funding code 01IW17001 (Deeplee).

\bibliography{acl2020}
\bibliographystyle{acl_natbib}

\end{document}